%% file: main.tex
\documentclass[10pt,twocolumn,letterpaper]{article}

\usepackage{iccv}
\usepackage{times}
\usepackage{epsfig}
\usepackage{graphicx}
\usepackage{ams math}
\usepackage{amssymb}

\usepackage[utf8]{inputenc} 
\usepackage[T1]{fontenc}    
\usepackage{url}            
\usepackage{booktabs}       
\usepackage{amsfonts}       
\usepackage{nicefrac}       
\usepackage{microtype}      
\usepackage{xcolor}         
\usepackage{tabu}         
\usepackage{xspace,mfirstuc,tabulary}

\usepackage{multirow}
\usepackage{array, caption, floatrow, makecell, booktabs}
\usepackage{wrapfig}
\usepackage{floatrow}
\usepackage{comment}
\usepackage{footnote}
\usepackage{enumitem}
\usepackage{accents}

\usepackage{graphicx}
\usepackage{subfigure}
\usepackage{bbding}
\usepackage{ntheorem}
\usepackage{enumitem}
\usepackage{oplotsymbl}
\usepackage{multirow}
\usepackage{bbm}
\usepackage{adjustbox}

\usepackage{wrapfig}
\usepackage{comment}
\usepackage{enumitem}
\usepackage{accents}

\usepackage{tabu}

\usepackage{multirow,array}
\usepackage{color, colortbl}
\definecolor{Gray}{gray}{0.93}
\definecolor{lightgray}{rgb}{0.9, 0.9, 0.9}

\usepackage{booktabs}
\usepackage{pifont}

\usepackage[pagebackref=true,breaklinks=true,colorlinks,bookmarks=false]{hyperref}
\usepackage{shadowtext}
\usepackage{floatrow}
\newlength\savewidth
\newcommand\paperurl[1]{{\footnotesize{\color{blue}{\url{#1}}}}}

\usepackage[breaklinks=true,bookmarks=false]{hyperref}

\iccvfinalcopy 

\def\iccvPaperID{****} 

\ificcvfinal\fi

\begin{document}

\title{Grounded SAM: Assembling Open-World Models for Diverse Visual Tasks}
\author{%
    International Digital Economy Academy (IDEA) \& Community\\
    \small{Code \& Demo: \href{https://github.com/IDEA-Research/Grounded-Segment-Anything}{https://github.com/IDEA-Research/Grounded-Segment-Anything}}
}

\ificcvfinal\thispagestyle{empty}\fi


\input{sections/00_abstract}

%

\input{sections/01_intro_new}

\input{sections/04_related_work}
\input{sections/02_methods}

\input{sections/03_experiments}
\input{sections/05_conclusion}
\input{sections/06_contribution_and_acknowledgments}

{\small
\bibliographystyle{ieee_fullname}
\bibliography{egbib}
}

\end{document}

%% file: sections/00_abstract.tex
\twocolumn[{%
\renewcommand\twocolumn[1][]{#1}%
\maketitle
\begin{center}
    \centering
    \includegraphics[width=0.98\textwidth]{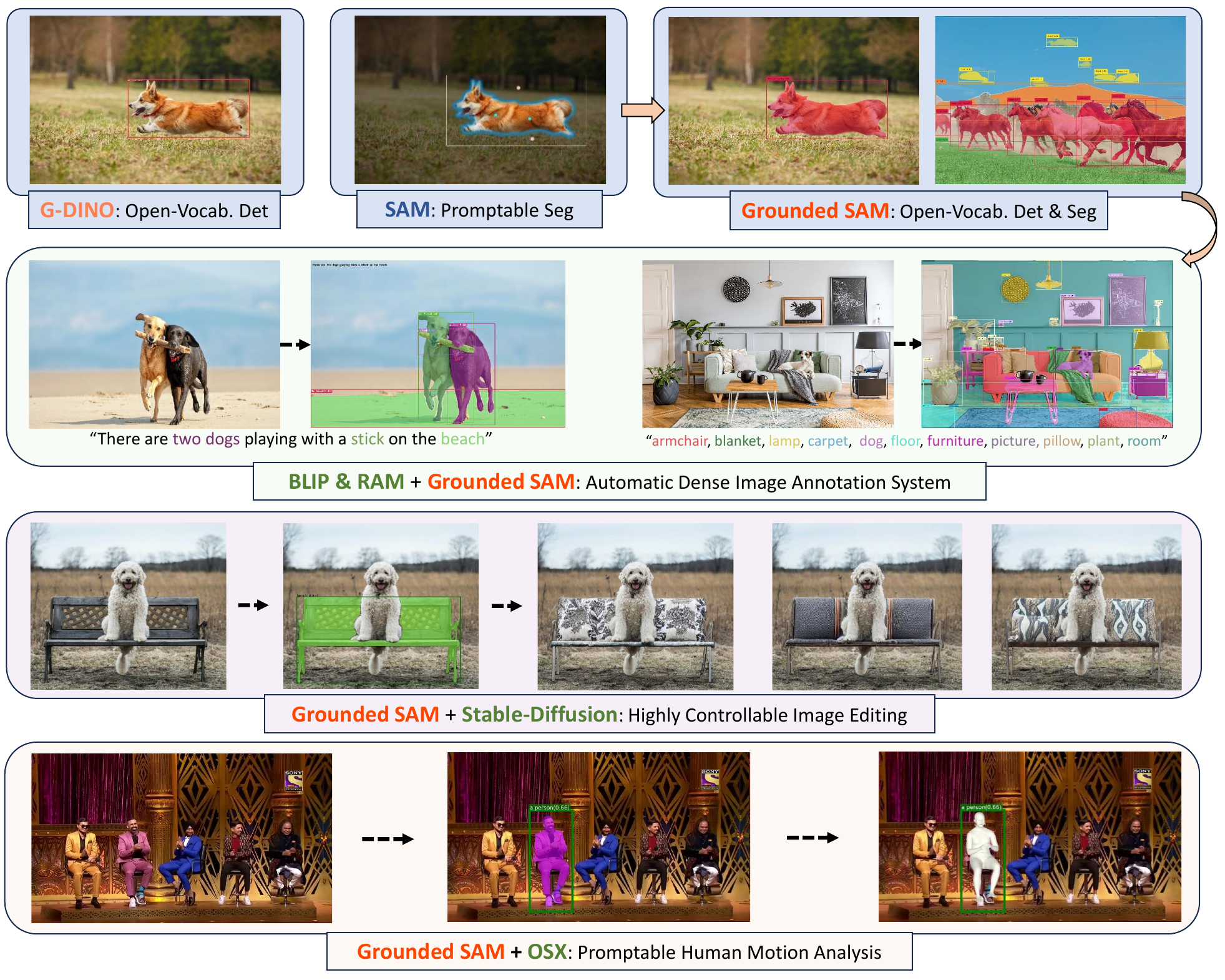}
\vspace{0.2em}
\captionof{figure}{\textbf{Grounded SAM} can simultaneously detect and segment corresponding regions within images based on arbitrary text inputs provided by users. And it can seamlessly integrate with other Open-World models to accomplish more intricate visual tasks
}
\label{fig:teaser}
\end{center}%
}]

\begin{abstract}
  We introduce \textbf{Grounded SAM}, which uses Grounding DINO~\cite{liu2023grounding} as an open-set object detector to combine with the segment anything model (SAM)~\cite{kirillov2023segment}. This integration enables the detection and segmentation of any regions based on arbitrary text inputs and opens a door to connecting various vision models. As shown in Fig.~\ref{fig:teaser}, a wide range of vision tasks can be achieved by using the versatile Grounded SAM pipeline. For example, an automatic annotation pipeline based solely on input images can be realized by incorporating models such as BLIP~\cite{li2022blip} and Recognize Anything~\cite{zhang2023recognize}. Additionally, incorporating Stable-Diffusion~\cite{rombach2022high} allows for controllable image editing, while the integration of OSX~\cite{lin2023osx} facilitates promptable 3D human motion analysis. Grounded SAM also shows superior performance on open-vocabulary benchmarks, achieving 48.7 mean AP on SegInW (Segmentation in the wild) zero-shot benchmark with the combination of Grounding DINO-Base and SAM-Huge models. 

\end{abstract}

%% file: sections/01_intro_new.tex
\section{Introduction}



Visual perception and understanding tasks in open-world scenarios are crucial for the advancement of applications such as autonomous driving, robotic navigation, and intelligent security surveillance. These applications demand robust and versatile visual perception models capable of interpreting and interacting with open-world environments. 

Currently, there are three primary methodologies to address the challenges in open-world visual perception.
First, the \textbf{Unified Model} approach involves training models like UNINEXT~\cite{UNINEXT} and OFA~\cite{wang2022ofa} on multiple datasets to support various vision tasks. This method also includes training large language models on different visual question-answering datasets to unify tasks, like LLaVA~\cite{LLaVA}, InstructBLIP~\cite{instructblip}, Qwen-VL~\cite{bai2023qwenvl} and other MLLMs~\cite{CogVLM, luo2023cheap, zhang2023llavagrounding}. However, a significant limitation of such an approach is its limited scope in data, especially in complex tasks like open-set segmentation.
Second, the \textbf{LLM as Controller} method attempts to bridge vision experts with language models. Examples include HuggingGPT~\cite{shen2023hugginggpt}, Visual ChatGPT~\cite{visualchatgpt}, and LLaVA-Plus~\cite{liu2023llavaplus}. These approaches leverage the linguistic comprehension capabilities of large language models to direct various visual tasks. However, this method is heavily reliant on the functionalities and limitations of large language models.
Third, the \textbf{Ensemble Foundation Models} approach seeks to accomplish open-world tasks in complex scenarios by collaboratively integrating expert models designed for specific contexts. This approach offers flexibility by combining the strengths of various specialized models.


Although there have been advances in addressing open-world tasks through these methodologies, a robust pipeline capable of supporting complex and fundamental open-world tasks such as open-set segmentation is still lacking in the market. Grounded SAM takes an innovative approach from the perspective of the Ensemble Foundation Models approach, pioneering the integration of open-set detector models, such as Grounding DINO~\cite{liu2023grounding}, and promptable segmentation models like SAM~\cite{kirillov2023segment}. It effectively tackles the open-set segmentation challenge by dividing it into two main components: open-set detection, and promptable segmentation. Based on this approach, Grounded SAM offers a powerful and comprehensive platform that further facilitates an efficient fusion of different expert models to tackle more intricate open-world tasks.



Building upon Grounded SAM as a foundation and leveraging its robust open-set segmentation capabilities, we can easily incorporating additional open-world models. For instance, when combined with Recognize Anything (RAM)~\cite{zhang2023recognize}, the RAM-Grounded-SAM model can automatically identify and segment things or objects within images without the need for any textual input, thus facilitating automatic image annotation tasks. Similar automatic image annotation capabilities can also be achieved through integration with BLIP~\cite{li2022blip}. Furthermore, Grounded SAM, when coupled with the inpainting capability of Stable Diffusion, as exemplified by the Grounded-SAM-SD model, can execute highly precise image editing tasks. We will provide a more detailed discussion of Grounded SAM and its augmented capabilities through the incorporation of additional open-world models in Section \ref{sec:playground}.

%% file: sections/04_related_work.tex
\section{Related Work}
\subsection{Task-specific Vision Models}

In the field of computer vision, significant advancements have been made across a variety of tasks, including image recognition~\cite{radford2021learning, li2022blip, Tag2Text, zhang2023recognize, huang2023openset}, generic object detection~\cite{ren2015faster,zhu2020deformable, meng2021conditional, li2022dn,zhang2022dino,liu2022dabdetr, jia2022detrs, liu2023grounding, ren2023strong, ren2023detrex, TRex, DFA3D}, generic image segmentation~\cite{cheng2021maskformer, BowenCheng2022Mask2FormerFV, DINOv,li2023maskdino,zhang2023mp,zou2022xdecoder, zhang2023simple, kirillov2023segment,zhang2023simple, hu2023segment, li2023semanticsam}, referring object detection and segmentation~\cite{luo2020multitask, liu2022dqdetr, SeqTR}, object tracking~\cite{yan2023bridging, zhang2022bytetrack}, image generation~\cite{Parti, Imagen, DALLE2, GLIDE, GigaGAN, Make-a-scene, rombach2022high,ju2023humansd,ControlNet,T2I-Adapter}, image editing~\cite{SDEdit, blended, Blend-diffusion, DreamBooth, ju2023direct}, human-centric perception and understanding~\cite{yang2023explicit, click_pose,yang2023unipose,yang2023semantic,yang2023boosting,cai2023smpler,lin2023osx,yang2023effective}, and human-centric motion generation~\cite{lu2023humantomato, chen2023humanmac,lin2023motion,wang2023physhoi,chen2024diffsheg}. However, despite these advancements, current models are mostly task-specific and usually fall short in addressing a broader range of tasks.


\subsection{Unified Models}

Unified models have been developed to address multiple tasks.
In the language field, Large language models (LLMs) such as GPT-3~\cite{floridi2020gpt}, LaMDA~\cite{thoppilan2022lamda}, and PaLM~\cite{chowdhery2022palm} are examples of general-purpose unified models, which handle language tasks through an auto-regressive and generative approach. 
Unlike language tasks that rely on a unified and structured token representation, vision tasks encompass many data formats, including pixel, spatial (e.g., box, key point), temporal, and others. 
Recent works have attempted to develop unified vision models from two perspectives to accommodate these diverse modalities. 
First, some models aim to unify various vision modalities into a single one. For instance, Pix2Seq~\cite{chen2021pix2seq} and OFA~\cite{wang2022ofa} attempt to merge spatial modalities such as box coordinates into language. 
Second, some models seek a unified model compatible with different modality outputs. UNINEXT~\cite{UNINEXT} is an example that supports different instance-level task outputs. 
Although these unified vision models are advancing the progress of general intelligence, existing models can only handle a limited number of tasks and often fall short of task-specific models in performance.

\subsection{Model Assembly with a Controller System}
Orthogonal to our work, {Visual ChatGPT}~\cite{visualchatgpt} and {HuggingGPT}~\cite{shen2023hugginggpt} propose to leverage LLMs to control different AI models for solving different tasks. 
Compared with these models, the foundation model assembling method does not employ an LLM as the controller, which makes the whole pipeline more efficient and flexible. We show that complex tasks can be decoupled, and step-by-step visual reasoning can be accomplished in a training-free model assembly manner.

%% file: sections/02_methods.tex
\section{Grounded SAM Playground}
\label{sec:playground}
In this chapter, utilizing Grounded SAM as a foundation, we demonstrate our method of amalgamating expert models from various domains to facilitate the accomplishment of more comprehensive visual tasks.

\subsection{Preliminary}
We discuss the basic components of Grounded SAM and other domain expert models here.

\textbf{Segment Anything Model (SAM)} \cite{kirillov2023segment} is an open-world segmentation model that can "cut out" any object in any image with proper prompts, like points, boxes, or text. It has been trained on over $11$ million images and $1.1$ billion masks. Despite of its strong zero-shot performance, the model cannot identify the masked objects based an arbitrary text input and normally requires point or box prompts to run.

\textbf{Grounding DINO} \cite{liu2023grounding} is an open-set object detector that can detect any objects with respect to an arbitrary free-form text prompt. The model was trained on over $10$ million images, including detection data, visual grounding data, and image-text pairs. It has a strong zero-shot detection performance. However, the model needs text as inputs and can only detect boxes with corresponding phrases.

\textbf{OSX} \cite{lin2023osx} is the state-of-the-art model for expressive whole-body mesh recovery, which aims to estimate the 3D human body poses, hand gestures, and facial expressions jointly from monocular images. It needs first to detect human boxes, crop and resize the human boxes, and then conduct single-person mesh recovery. 

\textbf{BLIP} \cite{li2022blip} is a vision-language model that unifies vision-language understanding and generation tasks. We use the image caption model of BLIP in our experiments. The caption model can generate descriptions given any image. However, the model cannot perform object-level tasks, like detecting or segmenting objects.

\textbf{Recognize Anything Model (RAM)} \cite{zhang2023recognize} is a strong image tagging model that can recognize any common categories of high accuracy for an input image. However, RAM can only generate tags but cannot generate precise boxes and masks for the recognized categories.

\textbf{Stable Diffusion} \cite{rombach2022high} is an image generation model that samples images from the learned distribution of training data. Its most widely used application is generating images with text prompts. We use its inpainting variant in our experiment. Despite of its awesome generation results, the model cannot perform perception or understanding tasks.


\textbf{ChatGPT \& GPT-4} \cite{gozalo2023chatgpt, openai2023gpt4} are large language models developed using the GPT (Generative Pre-trained Transformer) architecture, which is used for building conversational AI agents. It is trained on massive amounts of text data and can generate human-like responses to user input. The model can understand the context of the conversation and generate appropriate responses that are often indistinguishable from those of a human. 

\input{images_tex/grounded_sam_demo_images}

\subsection{Grounded SAM: Open-Vocabulary Detection and Segmentation} 
\label{sec:sec_auto_anno}

It is highly challenging to determining masks in images corresponding to regions mentioned in any user-provided text and thereby enabling finer-grained image understanding tasks like open-set segmentation. This is primarily due to the limited availability of high-quality data for segmentation in the wild tasks, which presents a challenge for the model to accomplish precise open-set segmentation under conditions characterized by data scarcity. In contrast, open-set detection tasks are more tractable, primarily due to the following two reasons. First, the annotation cost of detection data is relatively lower compared to segmentation tasks, enabling the collection of more higher-quality annotated data. Second, open-set detection only requires identifying the corresponding object coordinates on the images based on the given text without the need for precise pixel-level object masks.
Similarly, the prediction of the corresponding object mask, conditioned on a box and benefiting from the prior knowledge of the box's location, is more efficient than directly predicting the region mask based on a text. This approach has been validated in previous works such as OpenSeeD~\cite{zhang2023simple}, and the substantial issue of data scarcity can be largely addressed by utilizing the SAM-1B dataset developed in SAM~\cite{kirillov2023segment}.

Consequently, inspired by prior successful works such as Grounded Pre-training~\cite{zhang2022glipv2, liu2023grounding} and SAM~\cite{kirillov2023segment}, we aim to address complex segmentation in the wild tasks by combining the strong open-set foundation models. Given an input image and a text prompt, we first employ Grounding DINO to generate precise boxes for objects or regions within the image by leveraging the textual information as condition. Subsequently, the annotated boxes obtained through Grounding DINO serve as the box prompts for SAM to generate precise mask annotations. By leveraging the capabilities of these two robust expert models, the open-set detection and segmentation tasks can be more effortlessly accomplished. 
As illustrated in Fig.~\ref{fig:grounded_sam_demo_image}, Grounded SAM can accurately detect and segment text based on user input in both conventional and long-tail scenarios.

\subsection{RAM-Grounded-SAM: Automatic Dense Image Annotation}

The automatic image annotation system has numerous practical applications, such as enhancing the efficiency of manual annotation of data, reducing the cost of human annotation, or providing real-time scene annotation and understanding in autonomous driving to enhance driving safety. In the framework of Grounded SAM, it leverages the capabilities of Grounding DINO. Users have the flexibility to input arbitrary categories or captions, which are then automatically matched with entities within the images. Building upon this foundation, we can employ either an image-caption model (like BLIP~\cite{li2022blip} and Tag2Text~\cite{Tag2Text}) or an image tagging model (like RAM~\cite{zhang2023recognize}), using their output results (captions or tags) as inputs to Grounded SAM and generating precise box and mask for each instance. This enables the automatic labeling of an entire image, achieving an automated labeling system.
As depicted in Fig.~\ref{fig:ram_grounded_sam_demo_image}, RAM-Grounded-SAM exhibits the capability to automatically perform category prediction and provide dense annotations for input images across various scenarios. This significantly reduces the annotation cost and greatly enhances the flexibility of image annotation.

\input{images_tex/ram_grounded_sam_demo_images}

\subsection{Grounded-SAM-SD: Highly Accurate and Controllable Image Editing}
\label{sec:data_generation}

By integrating the powerful text-to-image capability of image generation models with Grounded SAM, we can establish a comprehensive framework that enables the creation of a robust data synthesis factory, supporting fine-grained operations at the part-level, instance-level, and semantic-level. As shown in Fig.~\ref{fig:grounded_sam_sd_demo_images}, users can obtain precise masks through interactive methods such as clicking or drawing bounding boxes within this pipeline. Moreover, users can leverage the capability of grounding, combined with text prompts, to automatically locate corresponding regions of interest. Building upon this foundation, with the additional capability of an image generation model, we can achieve highly precise and controlled image manipulation, including modifying the image representation, replacing objects, removing the corresponding regions, etc. In downstream scenarios where data scarcity arises, our system can generate new data, addressing the data requirements for the training of models. 

\input{images_tex/grounded_sam_sd_demo_images}

\subsection{Grounded-SAM-OSX: Promptable Human Motion Analysis}
\label{sec:understanding}

Previous expressive whole-body mesh recovery first detects \emph{all} (instance-agnostic) human boxes and then conducts the 
single-person mesh recovery. In many real-world applications, we need to specify the target person to be detected and analyzed. However, existing human detectors can not distinguish different instances (e.g., specify to analyze "a person with pink clothes"), making fine-grained human motion analysis challenging.
As shown in Fig.~\ref{fig:grounded_sam_osx_demo_image}, we can integrate the Grounded SAM and OSX~\cite{lin2023osx} models to achieve a novel promptable (instance-specific) whole-body human detection and mesh recovery, thereby realizing a promptable human motion analysis system.
Specifically, given an image and a prompt to refer to a specific person, we first use Grounded SAM to generate a precise specific human box. Then, we use OSX to estimate an instance-specific human mesh to complete the process.

\input{images_tex/grounded_sam_osx_demo_images}

\subsection{More Extensions for Grounded SAM}
\input{table_tex/seginw}

In addition to the aforementioned primary applications, Grounded SAM can further expand its scope of utilization by integrating more models. For instance, in the data labeling process, Grounded SAM can collaborate with the faster inference SAM models, such as FastSAM~\cite{fastsam}, MobileSAM~\cite{mobilesam}, Light-HQ-SAM~\cite{sam_hq}, and EfficientSAM~\cite{xiong2023efficientsam}. This collaboration can significantly reduce the overall inference time and expedite the labeling workflow. Grounded SAM can also leverage the HQ-SAM~\cite{sam_hq} model, which is capable of generating higher-quality masks, to enhance the quality of annotations. In the realm of image editing, Grounded SAM can also synergize with the newly proposed generative models such as Stable-Diffusion-XL~\cite{rombach2022high} to achieve higher-quality image editing. Furthermore, it can be integrated with models like LaMa~\cite{lama} and PaintByExample~\cite{paintbyexample} to achieve precise image erasure and customized image editing. Grounded SAM can also integrate with tracking models such as DEVA~\cite{deva_tracking} to perform object tracking based on specific text prompts.

%% file: images_tex/grounded_sam_demo_images.tex
\begin{figure*}[h]
    \centering
    \includegraphics[width=.98\linewidth]{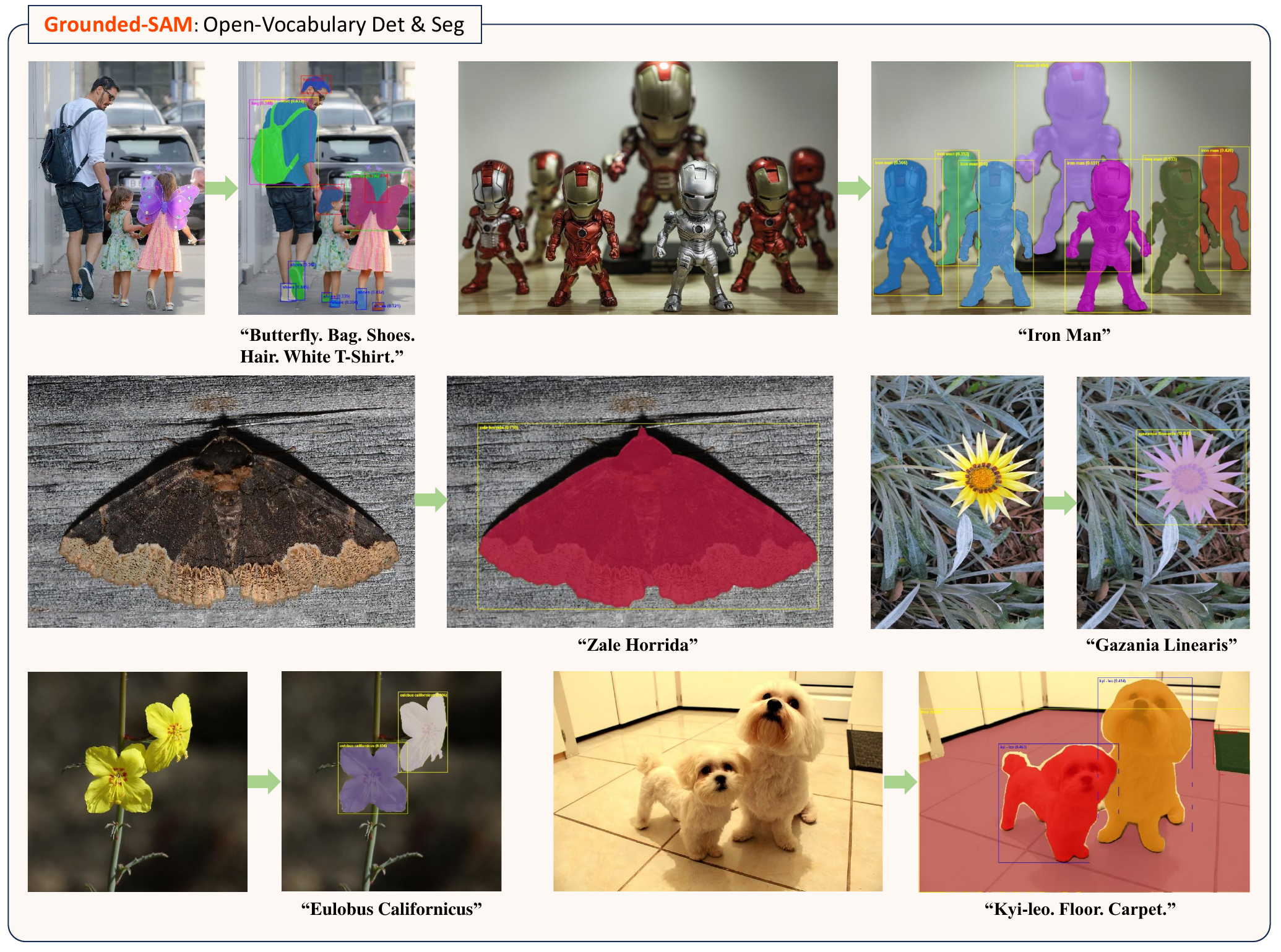}
    \caption{\textbf{Grounded-SAM} effectively detects and segments objects according to various user inputs. Its effectiveness is not limited to common cases but also includes long-tail object categories (like "Zale Horrida", and "Gazania Linearis", e.g.). Some of the demo images were sampled from the V3Det~\cite{wang2023v3det} dataset. We greatly appreciate their excellent work.}
    \label{fig:grounded_sam_demo_image}
\end{figure*}

%% file: images_tex/ram_grounded_sam_demo_images.tex
\begin{figure*}[h]
    \centering
    \includegraphics[width=.98\linewidth]{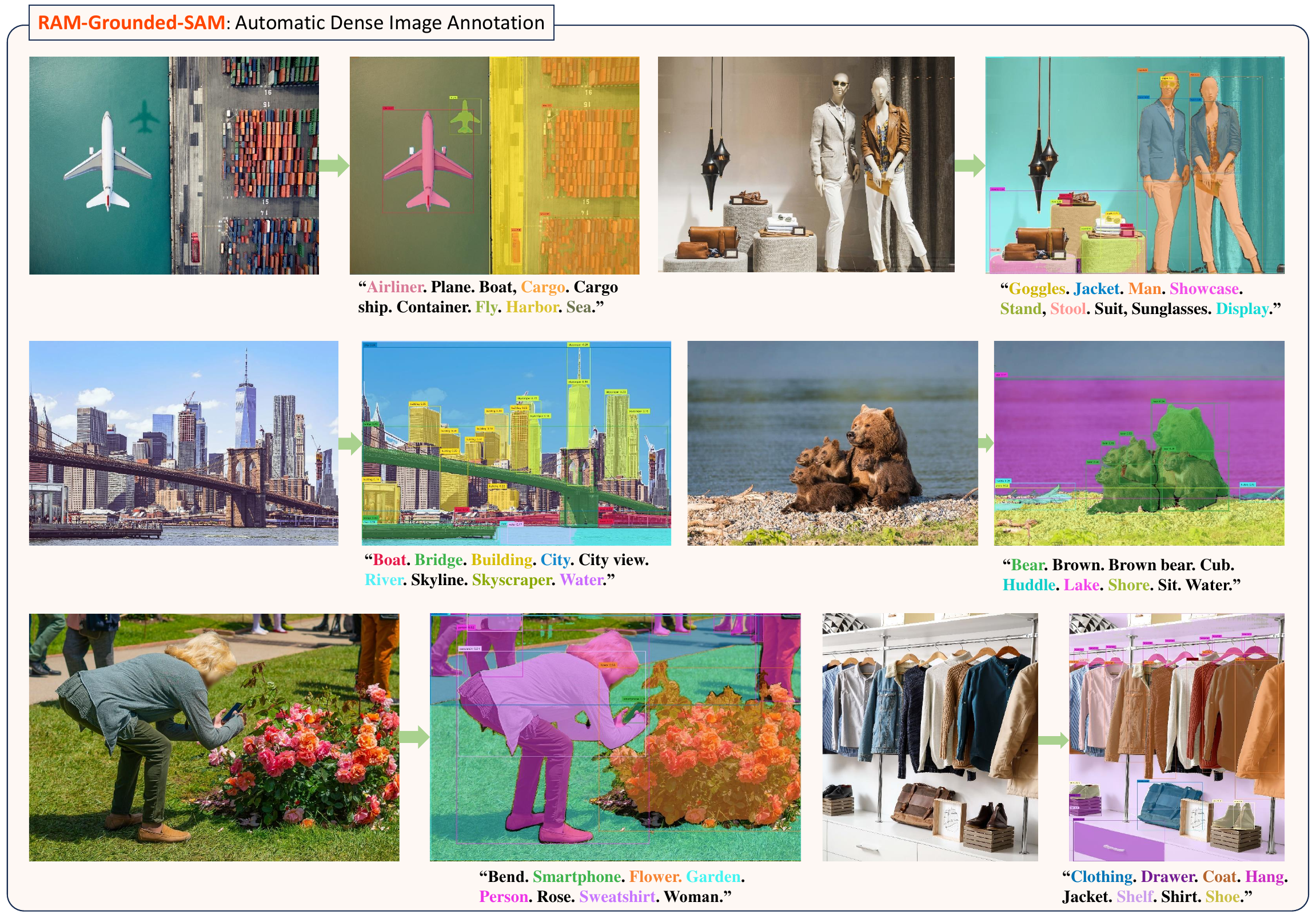}
    \caption{\textbf{RAM-Grounded-SAM} combines the robust tagging capabilities of the RAM~\cite{zhang2023recognize} with the open-set detection and segmentation abilities of Grounded SAM, which enables automatic dense image annotation with only image input (the demo images are sampled from the SA-1B~\cite{kirillov2023segment} dataset).}
    \label{fig:ram_grounded_sam_demo_image}
\end{figure*}

%% file: images_tex/grounded_sam_sd_demo_images.tex
\begin{figure*}[h]
    \centering
    \includegraphics[width=.98\linewidth]{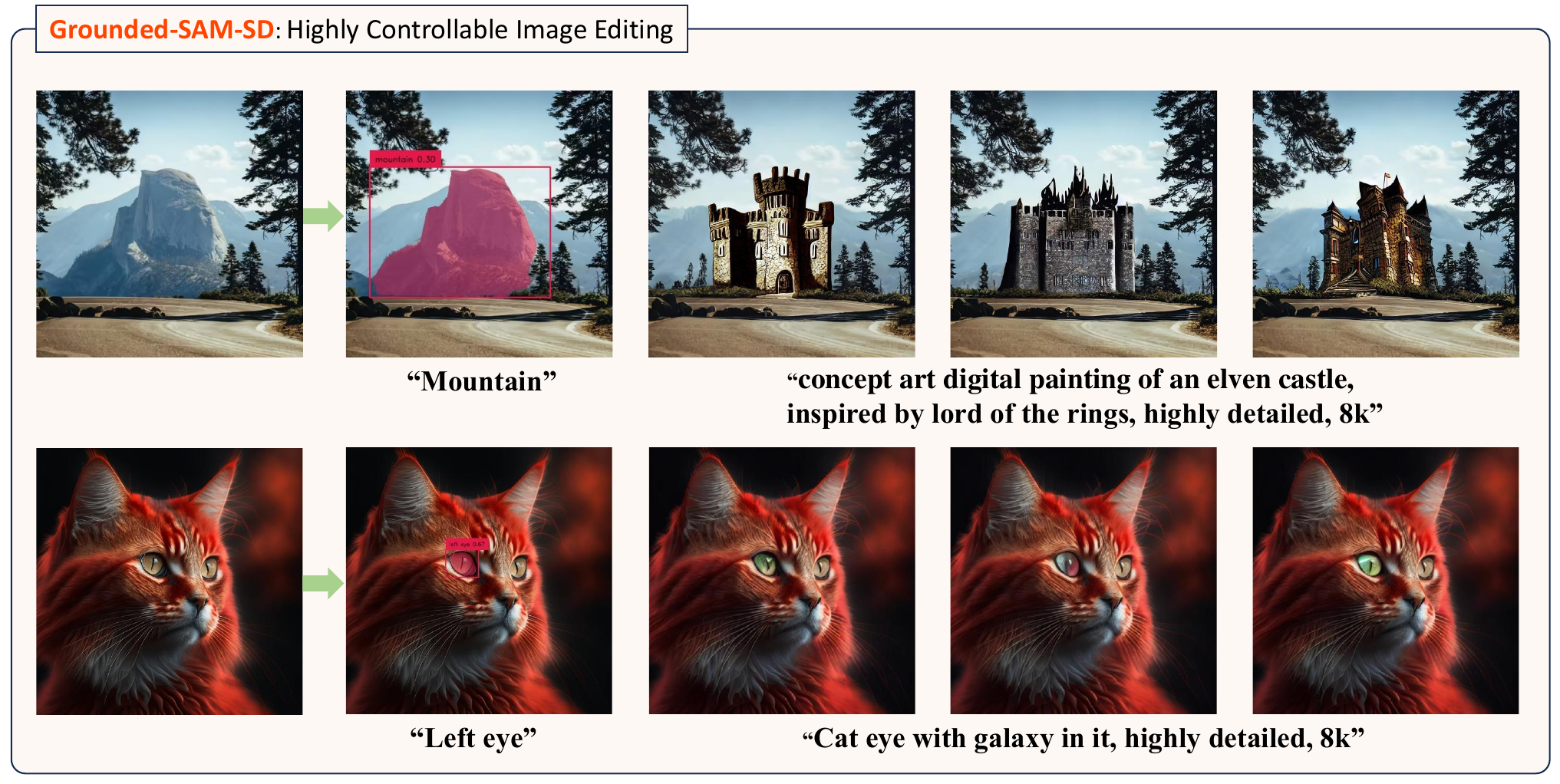}
    \caption{\textbf{Grounded-SAM-SD} combines the open-set ability of Grounded SAM with  inpainting }
    \label{fig:grounded_sam_sd_demo_images}
\end{figure*}

%% file: images_tex/grounded_sam_osx_demo_images.tex
\begin{figure*}[h]
    \centering
    \includegraphics[width=.98\linewidth]{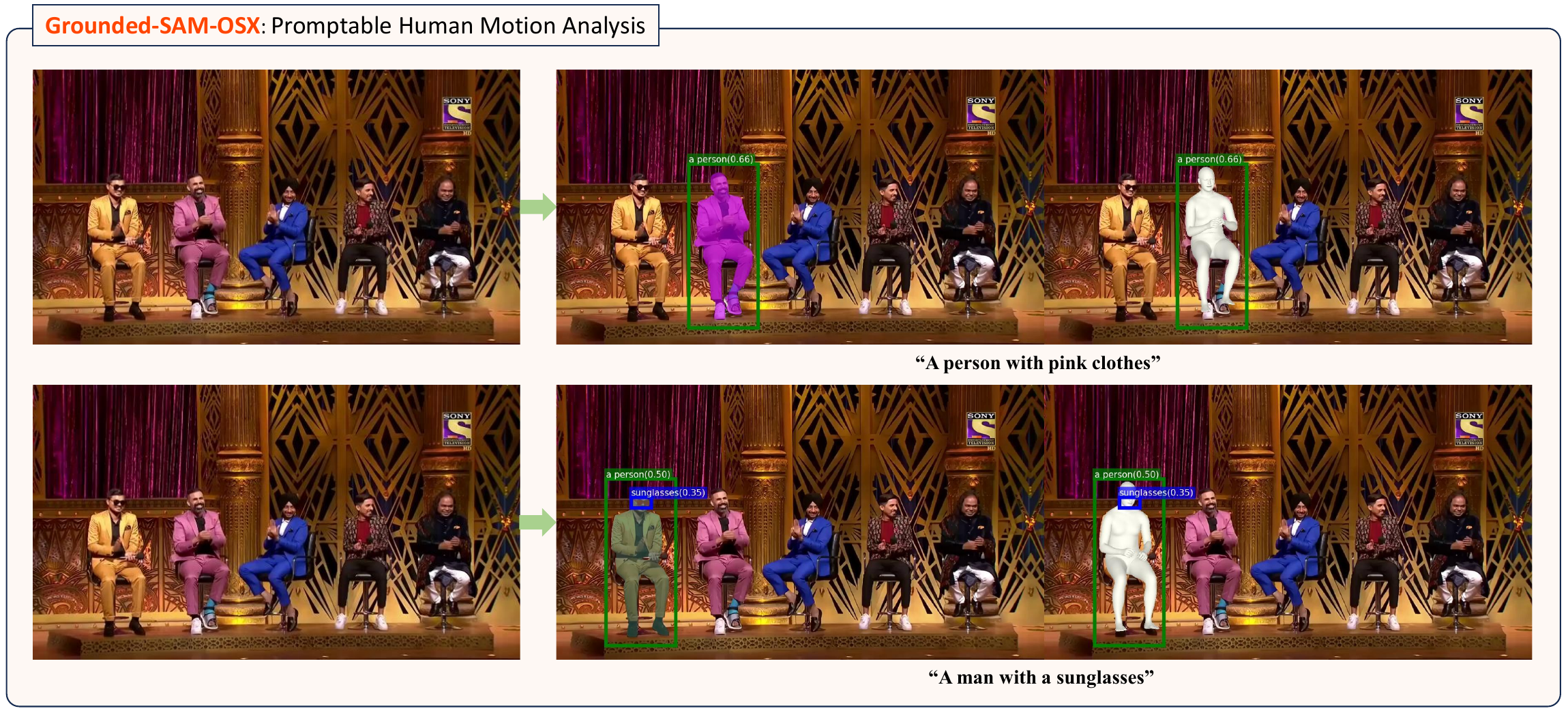}
    \caption{\textbf{Grounded-SAM-OSX} merges the text-promptable capability of Grounded SAM with the whole body mesh recovery ability of OSX~\cite{lin2023osx}, facilitating a precise human motion analysis system.}
    \label{fig:grounded_sam_osx_demo_image}
\end{figure*}

%% file: table_tex/seginw.tex
\begin{table*}[!h]
\renewcommand{\arraystretch}{1.5}
\caption{Zero-shot benchmarking results of Grounded-SAM in SGinW. The best and second-best results are highlighted in bold and underlined, respectively. * means the results were tested by the SAM-HQ~\cite{sam_hq} team. We are immensely thankful for their assistance in conducting these tests and highly appreciate their work.}
\label{table:sginw}
\begin{center}
\resizebox{1.0\linewidth}{!}{
\begin{tabular}{l|c|ccccccccccccccccccccccccc}
\toprule
Method & \rotatebox{90}{mean SGinW} & \rotatebox{90}{Elephants} & \rotatebox{90}{Hand-Metal} & \rotatebox{90}{Watermelon} & \rotatebox{90}{House-Parts} & \rotatebox{90}{HouseHold-Items} & \rotatebox{90}{Strawberry} & \rotatebox{90}{Fruits} & \rotatebox{90}{Nutterfly-Squireel} & \rotatebox{90}{Hand} & \rotatebox{90}{Garbage} & \rotatebox{90}{Chicken} & \rotatebox{90}{Rail} & \rotatebox{90}{Airplane-Parts} & \rotatebox{90}{Brain-Tumor} & \rotatebox{90}{Poles} & \rotatebox{90}{Electric-Shaver} & \rotatebox{90}{Bottles} & \rotatebox{90}{Toolkits} & \rotatebox{90}{Trash} & \rotatebox{90}{Salmon-Fillet} & \rotatebox{90}{Puppies} & \rotatebox{90}{Tablets} & \rotatebox{90}{Phones} & \rotatebox{90}{Cows} & \rotatebox{90}{Ginger-Garlic} \\
\hline
X-Decoder-T~\cite{zou2022xdecoder} & 22.6 & 65.6 & 22.4 & 16.2 & 5.5 & 50.6 & 41.6 & 66.5 & 62.1 & 0.6 & 28.7 & 12.0 & 0.7 & 10.5 & 1.1 & 3.6 & 1.2 & 19.0 & 9.5 & 19.3 & 15.0 & 48.9 & 15.2 & 29.9 & 12.0 & 7.9 \\
X-Decoder-L-IN22K & 26.6 & 63.9 & 20.3 & 13.5 & 4.9 & 50.5 & 74.4 & 79.1 & 58.8 & 0.0 & 24.3 & 3.5 & 1.3 & 12.3 & 0.5 & 13.4 & 18.8 & 43.2 & 14.6 & 20.1 & 12.3 & 57.3 & 6.9 & \textbf{43.4} & 12.3 & 15.6 \\
X-Decoder-B  & 27.7 & 68.0 & 18.5 & 13.0 & 6.7 & 51.7 & 81.6 & 76.7 & 53.1 & 20.6 & 30.2 & 13.6 & 0.8 & 13.0 & 0.3 & 5.6 & 4.2 & 45.9 & 13.9 & 27.3 & 18.2 & 55.4 & 8.0 & 8.9 & 36.8 & 19.4 \\
X-Decoder-L & 32.2 & 66.0 & 42.1 & 13.8 & 7.0 & 53.0 & 67.1 & 79.2 & 68.4 & 75.9 & 33.0 & 8.6 & 2.3 & 13.1 & 2.2 & \underline{20.1} & 7.5 & 42.1 & 9.9 & 22.3 & 19.0 & 59.0 & 22.5 & 15.6 & 44.9 & 11.6 \\
OpenSeeD-L~\cite{zhang2023simple} & 36.7 & 72.9 & 38.7 & 52.3 & 1.8 & 50.0 & 82.8 & 76.4 & 40.0 & \underline{92.7} & 16.9 & 82.9 & 1.8 & 13.0 & 2.1 & 4.6 & 4.7 & 39.7 & 15.4 & 15.3 & 15.0 & \textbf{74.6} & \textbf{47.4} & 7.6 & 40.9 & 13.6 \\
ODISE-L~\cite{xu2023odise} & 38.7 & 74.9 & 51.4 & 37.5 & \textbf{9.3} & \textbf{60.4} & 79.9 & 81.3 & 71.9 & 41.4 & \underline{39.8} & 84.1 & 2.8 & 15.8 & 2.9 & 0.4 & 18.3 & 37.7 & 15.0 & 28.6 & 30.2 & \underline{65.4} & 9.1 & \underline{43.8} & 41.6 & 23.0 \\
SAN-CLIP-ViT-L~\cite{xu2023side} & 41.4 & 67.4 & 62.9 & 43.5 & \underline{9.0} & \underline{60.1} & 81.8 & 77.4 & \underline{82.2} & 88.8 & \textbf{46.5} & 69.2 & 2.9 & 13.2 & 2.6 & 1.8 & 11.4 & 48.8 & \textbf{31.2} & \textbf{41.4} & 20.0 & 60.1 & 35.1 & 10.4 & 44.0 & 23.3 \\
UNINEXT-H~\cite{UNINEXT} & 42.1 & 72.1 & 57.0 & 56.3 & 0.0 & 54.0 & 80.7 & 81.1 & \textbf{84.1} & \textbf{93.7} & 16.9 & 75.2 & 0.0 & 15.1 & 2.6 & 13.4 & 71.2 & 46.1 & 10.1 & 10.8 & \textbf{44.4} & 64.6 & 21.0 & 6.1 & \textbf{52.7} & 23.7 \\
Grounded-HQ-SAM (B+H)*~\cite{sam_hq} & \textbf{49.6} & 77.5 & \textbf{81.2} & \textbf{65.6} & 8.5 & \underline{60.1} & \textbf{85.6} & \underline{82.3} & 77.1 & 74.8 & 25.0 & \underline{84.5} & \underline{7.7} & \underline{37.6} & \textbf{12.0} & \underline{20.1} & \textbf{72.1} & \textbf{66.3} & 21.8 & \underline{30.0} & \underline{42.2} & 50.1 & 29.7 & 35.3 & 47.8 & \underline{45.6} \\
\hline
\rowcolor{Gray}
Grounded-SAM (B+H)*  & \underline{48.7} & \underline{77.9} & \textbf{81.2} & \underline{64.2} & 8.4 & \underline{60.1} & \underline{83.5} & \underline{82.3} & 71.3 & 70.0 & 24.0 & \underline{84.5} & \textbf{8.7} & 37.2 & \underline{11.9} & \textbf{23.3} & \underline{71.7} & \underline{65.4} & 20.8 & \underline{30.0} & 32.9 & 50.1 & 29.8 & 35.4 & 47.5 & \textbf{45.8} \\
\rowcolor{Gray}
Grounded-SAM (L+H)  & 46.0 & \textbf{78.6} & \underline{75.2} & 61.5 & 7.2 & 35.0 & 82.5 & \textbf{86.9} & 70.9 & 90.7 & 28.2 & \textbf{84.6} & 7.2 & \textbf{38.4} & 10.2 & 17.4 & 59.7 & 43.7 & \underline{26.9} & 22.4 & 27.1 & 63.2 & \underline{38.6} & 3.4 & \underline{49.4} & 40.0 \\
\bottomrule
\end{tabular}
}
\end{center}
\vspace{-6mm}
\end{table*}

%% file: sections/03_experiments.tex
\section{Effectiveness of Grounded SAM}

To validate the effectiveness of Grounded SAM, we evaluate its performance on the Segmentation in the Wild (SGinW) zero-shot benchmark, which comprises 25 zero-shot in-the-wild datasets. As demonstrated in Table.~\ref{table:sginw}, the combination of Grounding DINO Base and Large Model with SAM-Huge results in significant performance improvements in the zero-shot settings of SGinW, when compared to previously unified open-set segmentation models such as UNINEXT~\cite{UNINEXT} and OpenSeeD~\cite{zhang2023simple}. 
By incorporating HQ-SAM~\cite{sam_hq}, which is capable of generating masks of higher quality than SAM, Grounded-HQ-SAM achieves even further performance improvement on SGinW.

%% file: sections/05_conclusion.tex
\section{Conclusion and Prospects}
The strengths of our proposed Grounded SAM and its extensions, which utilize the assembly of diverse expert models to accomplish various vision tasks, can be summarized as follows. First, the capability boundaries of the models can be seamlessly expanded by assembling various expert models. Previously, we could do $n$ tasks with $n$ models. Now, we can do up to $2^n-1$ tasks with $n$ expert models considering all possible model combinations. We can decouple a complex task into several sub-tasks that are solved by currently available expert models. Second, the model assembling pipeline is more explainable by decomposing a task into several sub-tasks. We can observe the output of each step to obtain the reasoning process of the final results. Finally, by combining various expert models, we can investigate new areas of research and applications, potentially leading to innovative results and technological advances.

\noindent\textbf{Prospects:} A significant prospect of our methodology entails establishing a closed loop between annotation data and model training. Through the combination of expert models, substantial annotation costs can be saved. Moreover, the inclusion of human annotators at different stages facilitates the filtering or fine-tuning of inaccurate model predictions, thereby enhancing the quality of model annotations. The annotated data is then continually utilized to further train and improve the model. Another potential application of our method is to combine with Large Language Models (LLMs). Given our assembled models can do almost any computer vision (CV) tasks with various input and output modalities, especially language, it becomes straightforward for LLMs to invoke our API via language prompts to effectively execute CV tasks. Last but not least, the model can be used to generate new datasets bridging any pairs of modalities, especially when combined with generation models.


%% file: sections/06_contribution_and_acknowledgments.tex
\section{Contributions and Acknowledgments}
We would like to express our deepest gratitude to multiple
persons from the research community for their substantial support in the Grounded SAM project. We have listed the main participating roles in the Grounded SAM Project below. Within each role, contributions are equal and are listed in a randomized order. Ordering within
each role does not indicate the ordering of the contributions.

\renewcommand{\baselinestretch}{1.3} \normalsize

\begingroup
\linespread{1.5}

\noindent
{\textbf{Leads}}

\noindent
\href{https://rentainhe.github.io/}{\textbf{Tianhe Ren}}, Co-Lead, Grounded SAM \& Grounded-SAM-SD pipeline. 

\noindent
\href{https://lsl.zone/}{\textbf{Shilong Liu}}, Co-Lead, Grounded SAM pipeline and online demo.

\noindent
\href{https://ailingzeng.site/}{\textbf{Ailing Zeng}}, Co-Lead, Grounded-SAM-OSX pipeline and demo.

\noindent
\href{https://jinglin7.github.io/}{\textbf{Jin Ling}}, Co-Lead, Grounded-SAM-OSX pipeline and demo.

\noindent
\href{https://github.com/CiaoHe}{\textbf{He Cao}}, Co-Lead, Grounded-SAM-SD pipeline and Interactive SAM Editing pipeline.

\noindent
\href{https://scholar.google.com/citations?user=D4tLSbsAAAAJ&hl=zh-CN}{\textbf{Kunchang Li}}, Co-Lead, BLIP-Grounded-SAM pipeline and ChatBot.

\noindent
\href{https://github.com/tuofeilunhifi}{\textbf{Jiayu Chen}}, Co-Lead, Grounded SAM modelscope demo support and code optimization.

\noindent
\href{https://xinyu1205.github.io/}{\textbf{Xinyu Huang}}, Co-Lead, RAM-Grounded-SAM demo support.

\noindent
\href{https://github.com/fengxiuyaun}{\textbf{Feng Yan}}, Co-Lead, Grounded SAM with VISAM tracking demo.

\noindent
\href{https://yukangchen.com/}{\textbf{Yukang Chen}}, Co-Lead, 3D-Box via Segment Anything.

\noindent
{\textbf{Core Contributors}}

\noindent
\href{https://scholar.google.com/citations?user=U_cvvUwAAAAJ&hl=zh-CN}{\textbf{Zhaoyang Zeng}}

\noindent
\href{https://haozhang534.github.io/}{\textbf{Hao Zhang}}

\noindent
\href{https://fengli-ust.github.io/}{\textbf{Feng Li}}

\noindent
\href{https://yangjie-cv.github.io/}{\textbf{Jie Yang}}

\noindent
\href{https://scholar.google.com/citations?user=zdgHNmkAAAAJ&hl=zh-CN}{\textbf{Hongyang Li}}

\noindent
\href{https://mountchicken.github.io/}{\textbf{Qing Jiang}}

\noindent
\href{https://chenxwh.github.io/}{\textbf{Chenxi Whitehouse}}

\noindent
\href{https://github.com/nomorewzx}{\textbf{Zhenxuan Wang}}

\endgroup

\noindent
{\textbf{Overall Technical Leads}}

\noindent
\href{https://www.leizhang.org/}{\textbf{Lei Zhang}}

\renewcommand{\baselinestretch}{1.0} \normalsize
\noindent